\documentclass[journal]{IEEEtran}
\usepackage{amsmath,amsfonts}
\usepackage[ruled,linesnumbered]{algorithm2e}
\usepackage{array}
\usepackage[caption=false,font=normalsize,labelfont=sf,textfont=sf]{subfig}
\usepackage{textcomp}
\usepackage{stfloats}
\usepackage{afterpage}
\usepackage{url}
\usepackage{verbatim}
\usepackage{graphicx}
\usepackage{multirow}
\usepackage{tabularx}
\newlength{\Lcol}
\newlength{\Xcol}
\def\BibTeX{{\rm B\kern-.05em{\sc i\kern-.025em b}\kern-.08em
    T\kern-.1667em\lower.7ex\hbox{E}\kern-.125emX}}
\usepackage{balance}
\usepackage{xr}
\begin{document}
\sloppy
\emergencystretch=3em
\title{AgniNav: Configuration-Driven Cross-Embodiment Local Planning for Robot Navigation}
\author{Tianhao Zang$^{1,2\dagger}$, Cheng Siwei$^{3\dagger}$, Haidong Huang$^{2}$, Shanze Wang$^{1}$, Wei Zhang$^{1*}$%
\thanks{$^\dagger$These authors contributed equally to this work.}%
\thanks{$^1$College of Information Science and Technology, Eastern Institute of Technology, Ningbo, P. R. China.}%
\thanks{$^2$University of Nottingham, Nottingham, UK.}%
\thanks{$^3$University of Science and Technology of China, Hefei, P. R. China.}%
\thanks{$^*$Corresponding author. Email: zhw@eitech.edu.cn.}%
}

\markboth{IEEE/ASME Transactions on Mechatronics}%
{Zang \MakeLowercase{\textit{et al.}}: AgniNav: Configuration-Driven Cross-Embodiment Local Planning for Robot Navigation}

\maketitle

\begin{abstract}
Monocular local navigation is attractive for lightweight robots, but existing vision-based policies often couple perception to a specific body, camera height, and footprint, making transfer from wheeled bases to legged platforms dependent on retraining or active depth hardware. This paper introduces AgniNav, a configuration-driven local navigation framework that standardizes cross-embodiment transfer at the collision-envelope level. Each robot is specified by a measurable four-parameter safety envelope: collision-relevant height, front length, rear length, and half width. The height parameter conditions an image-to-scan network to predict a one-dimensional, collision-relevant pseudo-laserscan from a monocular color image, while the remaining footprint parameters configure a dimension-aware local planner for collision checking. Training uses height-conditioned column-minimum scan labels generated from paired color-depth data, allowing the same image to supervise different safety envelopes without collecting robot-specific data. To the best of our knowledge, AgniNav is the first monocular local-navigation framework that jointly conditions perception and planning on a shared collision-envelope configuration for zero-retraining deployment across wheeled, quadruped, and humanoid platforms. Real-robot experiments on a Turtlebot2, Unitree Go2, and Accelerated Evolution K1 achieve 39/40, 18/20, and 18/20 successes with 0/40, 1/20, and 2/20 collisions, respectively, while running at 30~Hz on Jetson Orin.
\end{abstract}
\

\begin{IEEEkeywords}
Vision-Based Navigation, Collision Avoidance, Deep Learning for Visual Perception, Cross-Embodiment Navigation, Representation Learning.
\end{IEEEkeywords}

\section{Introduction}
\IEEEPARstart{M}{obile} robot navigation aims to identify a feasible, safe path to a destination within complex environments, a fundamental capability for applications ranging from autonomous logistics to human-centric service robots. While traditional local planning algorithms, such as the Dynamic Window Approach (DWA) and Timed Elastic Band (TEB), have proven highly effective, their reliance on precise spatial representations and dense 3D mapping often introduces significant computational overhead. Alternatively, mapless navigation approaches, particularly those driven by Deep Reinforcement Learning (DRL), directly process raw sensor observations to compute steering commands. This paradigm enables robots to navigate dynamic environments without explicit map dependence, showing great potential for scalable real-world deployment.

For reliable metric obstacle avoidance, standard local planners universally demand an explicit geometric boundary. The 1D metric laserscan has long served as the predominant interface for this task, offering a compact spatial representation with explicit safety margins while enabling dimension-configurable policies to adapt to varying physical footprints. However, physical LiDARs are typically expensive and structurally prominent. Conversely, monocular cameras offer a highly scalable, low-cost alternative with rich texture information \cite{xiao2022motion}, \cite{jiang2022autonomous}. Despite these advantages, achieving robust generalization across structurally diverse robots---spanning differential-drive wheels, quadrupeds, and humanoids---remains a critical challenge for monocular systems. Raw visual observations inherently suffer from scale ambiguity; projecting a 3D scene into a 2D image obliterates absolute metric scale. By bypassing the explicit 1D geometric boundary, vision-based end-to-end policies often tightly couple perception to a specific robot's kinematics, footprint, and exact camera mounting height \cite{kim2018end}, \cite{kulhanek2021visual}, \cite{gupta2017cognitive}. Minor alterations in camera pitch or transitions between disparate robot embodiments lead to severe domain shifts, requiring extensive and costly retraining.

To address these visual navigation challenges, methods that attempt to recover explicit geometry via cascading monocular depth estimation with dense 3D volumetric mapping overcomplicate the state space, introducing unacceptable latency for highly dynamic platforms \cite{yan20203d}, \cite{chakravarty2019gen}. Other learning-based methods focus on massive, diverse datasets or domain fine-tuning to learn generalizable visual representations. However, these methods rarely incorporate the robot's explicit geometric footprint and kinematic constraints, thereby introducing significant safety risks during local planning in tight spaces. Standard local planners require reliable 1D obstacle representations \cite{brock1999high}, \cite{ROSMANN2017142}, a requirement that current monocular approaches struggle to bridge efficiently without sacrificing frame rates or cross-platform generalizability.

To overcome these limitations, AgniNav asks: How can a monocular system produce the 1D metric boundary required by local planners while adapting to different robot bodies through measurable physical parameters? We represent each robot with a compact collision-relevant embodiment configuration $\mathbf{c}_e=[H_{\max}, L_1, L_2, W/2]$, where $H_{\max}$ describes the upper bound of the configured rigid collision envelope protected by perception, and $(L_1, L_2, W/2)$ describe the horizontal safety envelope used by the planner. This tuple is not intended as a complete morphology representation; instead, it is a local-navigation proxy for the robot's 2.5D collision envelope.

At the core of our system is image2scan (I2S), an efficient multi-task learning architecture that predicts a 1D pseudo-laserscan from a monocular RGB image conditioned on $H_{\max}$. Its scan labels are generated from dense depth maps using a height-conditioned column-minimum strategy (Section~IV-B). The resulting pseudo-scan is passed to a Deep Reinforcement Learning-Based Dimension-Configurable Local Planner (DRL-DCLP) \cite{10900448}, which consumes $(L_1, L_2, W/2)$ as footprint parameters. In this way, I2S and DRL-DCLP share one configuration interface: $H_{\max}$ shapes perception, while the remaining tuple entries shape planning.

\begin{figure*}[t]
\centering
\includegraphics[width=\textwidth]{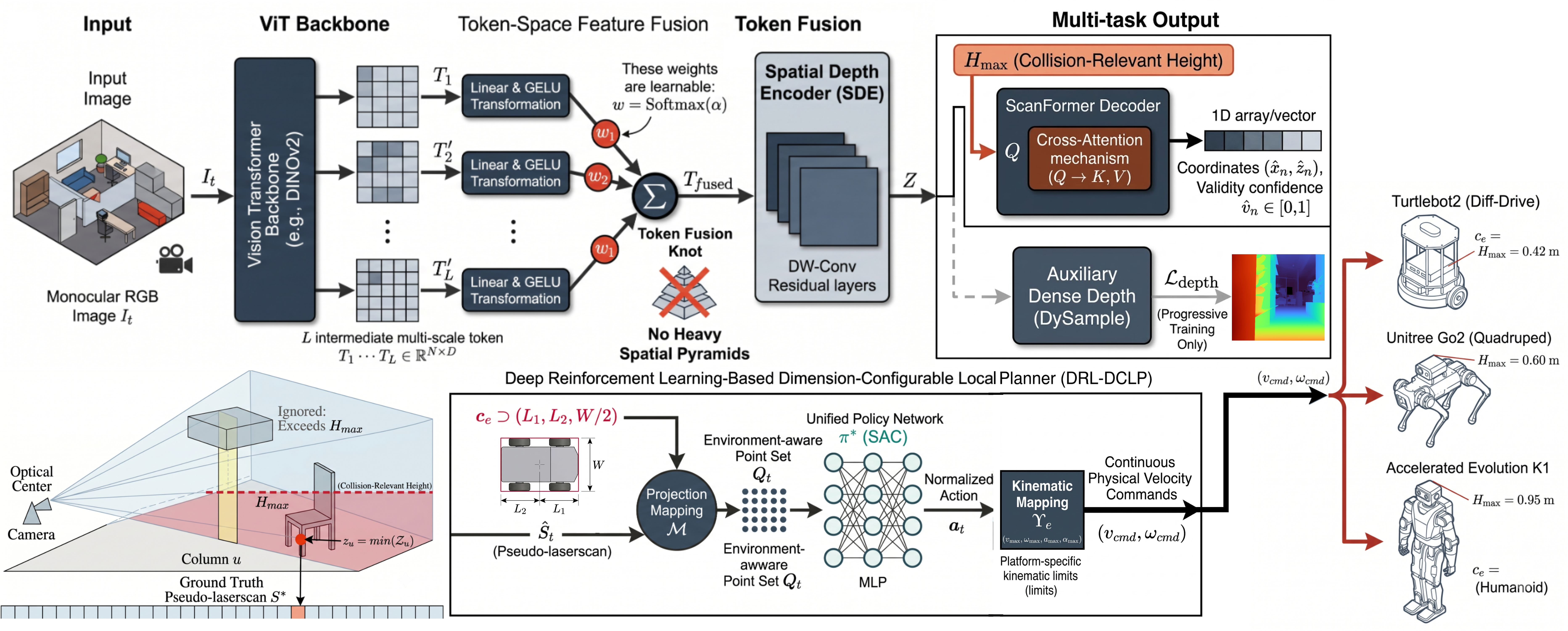}
\caption{\footnotesize Overall framework of AgniNav. A monocular RGB image is encoded by a ViT+SDT backbone, and ScanFormer decodes the features into a 1D pseudo-laserscan conditioned on $H_{\max}$. During training, geometry-derived scan pseudo-labels are generated from calibrated RGB-D data using a height-conditioned column-minimum strategy. At inference, the target robot is specified by $\mathbf{c}_e=[H_{\max},L_1,L_2,W/2]$: I2S predicts the scan from $(I_t, H_{\max})$, and DRL-DCLP uses $(L_1,L_2,W/2)$ for footprint-aware local planning.}
\label{fig:agninav_overview}
\end{figure*}

We extensively validate AgniNav in diverse real-world environments, including teaching buildings, corridors, offices, and laboratories. As illustrated in Fig.~\ref{fig:exp_traj_gallery}, AgniNav is successfully deployed across three embodiment-diverse platforms: a Turtlebot2 (differential-drive), a Unitree Go2 (quadruped), and an Accelerated Evolution K1 (humanoid). Compared to traditional monocular depth-and-mapping baselines, our streamlined approach substantially reduces computational overhead, achieving a real-time inference rate of over 30 Hz onboard resource-constrained edge computers, such as the NVIDIA Jetson Orin (FP16), widely used in modern robotic platforms. The primary contributions of this paper are threefold:
\begin{itemize}
\item We propose AgniNav, a configuration-driven cross-platform local planning framework for collision-envelope-level embodiment transfer. The method does not attempt to model full-body morphology, contact dynamics, or legged locomotion intelligence; instead, it standardizes local obstacle avoidance through a measurable 2.5D safety-envelope configuration $\mathbf{c}_e=[H_{\max},L_1,L_2,W/2]$. I2S predicts a 1D pseudo-laserscan conditioned on $H_{\max}$, while DRL-DCLP uses $(L_1, L_2, W/2)$ for footprint-aware local planning.

\item We formalize a visibility-aware column-minimum strategy as the training-time label generation mechanism that enables configuration-conditioned perception. By generating geometry-derived scan pseudo-labels conditioned on $H_{\max}$ and training on a calibrated dataset spanning multiple $H_{\max}$ values, the network learns collision-relevant obstacle boundaries rather than a camera-height-specific depth-to-scan conversion.

\item We demonstrate that token-space feature fusion via the Simple Depth Transformer (SDT) achieves comparable geometric fidelity to dense spatial pyramids (DPT), providing a practical, resource-efficient alternative for edge deployment.
\end{itemize}
\section{Related Work}
Traditional mapless DRL navigation methods suffer from the coupling problem, where perception and action are inherently overfit to the training robot's camera extrinsics and footprint. Existing cross-modal methods often rely on unsafe center-row heuristics or computationally heavy spatial pyramids. AgniNav addresses this by decoupling perception from platform constraints using a lightweight Simple Depth Transformer and a continuous $H_{\max}$ conditioning interface. A comprehensive literature review and positioning against Foundation Models and SLAM pipelines (including Table 1) are provided in the supplementary material.

\section{Problem Formulation}

The primary objective of AgniNav is to enable safe local navigation with high success rates for diverse robotic embodiments using only monocular RGB observations. We decompose the navigation pipeline into two sequential tasks: cross-modal perception (predicting a standardized 2D laserscan from an image) and dimension-configurable local planning.

\subsection{Collision-Relevant Embodiment Configuration}

We define a compact collision-relevant embodiment configuration:
\begin{equation*}
\mathbf{c}_e = [H_{\max}, L_1, L_2, W/2],
\end{equation*}
where $H_{\max}$ is the upper height of the configured rigid collision envelope that must be protected by perception, $L_1$ and $L_2$ are the forward and rear safety-envelope lengths, and $W/2$ is the half-width used for footprint-aware collision checking. This configuration should not be interpreted as a complete morphology representation: it does not model mass, leg length, joint limits, gait, compliance, or full-body dynamics. Instead, $\mathbf{c}_e$ is a low-dimensional, physically measurable proxy for the robot's local-navigation collision envelope.

AgniNav uses $\mathbf{c}_e$ as the common interface between perception and planning:
\begin{equation*}
\hat{S}_t = f_{\theta}(I_t, H_{\max}), \qquad
a_t = \pi(R_t, L_1, L_2, W/2, \kappa_e),
\end{equation*}
where $R_t$ is the discretized pseudo-laserscan and $\kappa_e=(v_{max}, \omega_{max}, a_{max}, \alpha_{max})$ denotes platform-specific kinematic limits. Thus, $H_{\max}$ controls which vertical obstacles are collision-relevant during scan prediction, while $(L_1,L_2,W/2)$ controls the horizontal footprint used by the planner.

\subsection{Coordinate Systems and Pseudo-Scan Label Formalization}

Let $\mathcal{I} \subset \mathbb{R}^{H \times W \times 3}$ denote the space of monocular RGB images captured by the robot's forward-facing camera. At any discrete time step $t$, the robot receives an observation $I_t \in \mathcal{I}$. Traditional metric local planners require a 2D obstacle representation, typically acquired via a physical LiDAR sensor. We use a calibrated pinhole camera frame in which $x$ is lateral, $z$ is forward depth, and $y_c$ is the camera vertical coordinate that increases with image row index. Collision relevance is evaluated in the ground frame using the metric height $h$ above the floor. We formalize the training target as a geometry-derived 2D pseudo-scan $S^*_t = \{(x_i^*, z_i^*)\}_{i=1}^{W_s}$, where $x_i^*$ and $z_i^*$ represent the Cartesian horizontal coordinate and forward depth of the $i$-th nearest collision-relevant obstacle, respectively.

The derivation of this pseudo-scan from depth maps must address a fundamental geometric challenge: center-row extraction, which samples depths only along the camera's optical center plane, inherently overlooks obstacles distributed off the horizontal plane (Section~IV-B). Our proposed column-minimum strategy resolves this by evaluating the full vertical extent of each image column.

\subsection{MDP Formulation and Reward Design}
We formulate the dimension-configurable local planning problem as a discrete-time MDP. The state incorporates a continuous parameterization of the collision envelope $(L_1, L_2, W/2)$ directly alongside the discretized pseudo-laserscan and kinematic limits. The reward function provides a sparse terminal reward for success, a large penalty for collision, and dense progress shaping. The exact formal state representation, distance embedding equations, and piecewise reward formulations are detailed in the supplementary material.

\section{Methodology}
\subsection{System Overview}

We introduce \textit{image2scan} (I2S), a neural architecture that translates monocular RGB observations into a configuration-conditioned 1D pseudo-laserscan in a single forward pass. The network takes both the RGB image and $H_{\max}$ as inputs, with $H_{\max}$ concatenated as a scalar conditioning signal at the ScanFormer decoder (Section~IV-C). Training supervision is obtained from dense depth maps through the height-conditioned column-minimum strategy described below.

We define the perception network as a parameterized deep neural network $f_\theta: \mathcal{I} \times \mathbb{R} \rightarrow \mathcal{S}$, where $\mathcal{S} \subset \mathbb{R}^{W_s \times 3}$ is the target scan space. The network maps the RGB input and the height parameter to a 1D pseudo-laserscan:
\begin{equation}
\hat{S}_t = f_\theta(I_t, H_{\max}) = \{(\hat{x}_i, \hat{z}_i, \hat{v}_i)\}_{i=1}^{W_s}
\end{equation}
where $W_s$ is the fixed horizontal resolution of the scan (e.g., $W_s = 640$), $\hat{x}_i$ and $\hat{z}_i$ are the Cartesian coordinates of the $i$-th scan point, and $\hat{v}_i \in [0, 1]$ represents the validity confidence. The raw prediction is then filtered to yield a valid obstacle set:
\begin{equation}
\bar{S}_t = \{(\hat{x}_i, \hat{z}_i) \mid \hat{v}_i \ge \tau, \sqrt{\hat{x}_i^2 + \hat{z}_i^2} \in [r_{min}, r_{max}]\}
\end{equation}
where $\tau$ is the confidence threshold, and $[r_{min}, r_{max}]$ defines the valid sensing range. Finally, $\bar{S}_t$ is discretized into a uniform polar range array $R_t \in \mathbb{R}^K$ to interface directly with standard local planners.

\textbf{Runtime rectification and scan discretization.}
At inference, different robots obtain configuration-specific scans through the $H_{\max}$ conditioning input. For legged platforms, the camera image is first rectified to the nominal zero-pitch training frame using the synchronized IMU roll/pitch estimate and calibrated intrinsics. The rotation-only rectification uses the standard homography $H_{\mathrm{imu}}=K R_{\mathrm{imu}}^{-1}K^{-1}$, where $R_{\mathrm{imu}}$ is the measured camera attitude relative to the nominal camera frame; this step requires only the RGB image and IMU orientation, not an online depth map. The predicted Cartesian points are then interpreted in the rectified camera frame and transformed to the robot base frame using the calibrated static camera extrinsics before publication.

The mapping from the dense I2S output to the planner input is deterministic. Each scan query corresponds to a calibrated horizontal ray inside the camera field of view, with angle $\phi_i=\arctan(\hat{x}_i/\hat{z}_i)$ and range $r_i=\sqrt{\hat{x}_i^2+\hat{z}_i^2}$. Points failing the confidence or range filters in Eq.~(6) are discarded. For the ROS \texttt{sensor\_msgs/LaserScan} message, the angular limits are set to the calibrated camera horizontal field of view $(\phi_{\min},\phi_{\max})$, and the range array is filled by taking the minimum valid range falling into each angular cell; cells with no valid return are assigned $r_{\max}$. Before DRL-DCLP, the published scan is downsampled to $K=64$ uniform polar bins using the same nearest-obstacle rule, preserving the closest collision-relevant obstacle in each bin. The pseudo-scan is therefore not simply a depth-to-scan conversion: it is a projection of 3D space into the robot's collision-relevant navigation boundary with explicit attitude compensation and planner-compatible angular binning.

\subsubsection{$H_{\max}$ Conditioning Mechanism}
\label{sec:hmax_mechanism}
During scan-supervised training, $H_{\max}$ is not obtained by physically recollecting the same scene with different robots. Instead, we collect a calibrated robot RGB-D frame once and use the depth map, camera intrinsics, camera-to-ground extrinsics, and traversability mask to analytically generate multiple height-conditioned pseudo-scan labels. Specifically, each calibrated raw tuple $(I_i, D_i, K_i, T_i, M_i)$ is expanded into multiple supervised samples $\{(I_i, H_j), S^*_{i,j}\}$, where $H_j \sim \text{Uniform}(0.20\,\text{m}, 0.90\,\text{m})$. The RGB image $I_i$ is identical across these samples, while the target pseudo-laserscan $S^*_{i,j}$ changes because the column-minimum projection only considers obstacle points whose height above the ground is within the collision-relevant height $H_j$. Therefore, $H_{\max}$ acts as a supervised conditioning variable rather than passive metadata.

Formally, given the calibrated custom RGB-D dataset $\mathcal{D}_{cal} = \{(I_i, D_i, K_i, T_i, M_i)\}_{i=1}^N$, the conditioned scan-supervision dataset is defined as:
\begin{equation}
\begin{aligned}
\mathcal{D}_{scan} = \{&((I_i, H_j), S^*_{i,j}) \mid i = 1 \dots N, \\
& H_j \sim \text{Uniform}(0.20\,\text{m}, 0.90\,\text{m})\}
\end{aligned}
\end{equation}

The scalar $H_{\max}$ is concatenated as a conditioning signal to the ScanFormer queries (Section~IV-C), enabling different scans for the same image depending on the target platform's collision-relevant height. To simulate target camera perspectives, the custom scan-labeled dataset was collected using an Orbbec Femto Bolt active ToF RGB-D camera (providing ToF depth and color images) mounted on a single Turtlebot2 robot at three calibrated camera heights: $z_{cam} \in \{0.42\,\text{m}, 0.60\,\text{m}, 0.90\,\text{m}\}$. Note that the deployment camera setups (such as the WHEELTEC C100 in Table~\ref{tab:robot_configs}) are monocular color cameras, demonstrating our model's capability to generalize from active ToF-supervised training to monocular deployment. All scan labels are generated in the same nominal rectified camera frame used at deployment. Scan pseudo-labels are generated on-the-fly using the column-minimum strategy with randomly sampled $H_{\max}$ values. Each training batch uniformly samples camera-height perspectives and height labels. NYU Depth V2 is used only for the auxiliary dense-depth objective in the progressive curriculum; it is never used to generate height-conditioned scan labels because it lacks the robot-specific camera-to-ground extrinsics required by the height projection in Section~IV-B. In this paper, $H_{\max}$ denotes the robot's \textit{collision-relevant height}: the vertical extent of the configured rigid collision envelope that must be protected by the local planner. It is distinct from the manufacturer's nominal full-body height and may include elevated sensing fixtures when they are part of the collision envelope. For DMR1--DMR3, the Turtlebot2 base is lower than the camera mount, but the camera support is included in collision checking; therefore, $H_{\max}$ is set to $0.42$~m to match the protected vertical envelope. The same rule is applied to DMR4: the quadruped body and rigid camera/support assembly are included in the protected envelope, yielding $H_{\max}=0.60$~m.

\begin{table*}[!t]
\renewcommand{\arraystretch}{1.3}
\caption{Collision-Relevant Embodiment Configurations for Real-World Cross-Embodiment Experiments}
\label{tab:robot_configs}
\centering
\small
\begin{tabular}{llcccccccc}
\hline
\textbf{Config ID} & \textbf{Platform Type} & \multicolumn{4}{c}{\textbf{$\mathbf{c}_e$ (m)}} & \multicolumn{2}{c}{\textbf{Sensor Extrinsics (m)}} & \textbf{Camera Setup} & $v_{\max}$ \textbf{(m/s)} \\
\cline{3-6} \cline{7-8}
& & $H_{\max}$ & $L_1$ & $L_2$ & $W/2$ & $z_{\text{cam}}$ & $x_{\text{cam}}$ & & \\
\hline
\textbf{DMR1} (Base) & Diff-Drive & 0.42 & 0.17 & 0.17 & 0.18 & 0.42 & +0.03 & $1 \times$ WHEELTEC C100 & 0.5 \\
\textbf{DMR2} (Long) & Diff-Drive & 0.42 & 0.285 & 0.285 & 0.18 & 0.42 & +0.03 & $1 \times$ WHEELTEC C100 & 0.5 \\
\textbf{DMR3} (Wide) & Diff-Drive & 0.42 & 0.17 & 0.17 & 0.40 & 0.42 & +0.03 & $1 \times$ WHEELTEC C100 & 0.5 \\
\hline
\textbf{DMR4} (Quad) & Quadruped  & 0.60 & 0.35 & 0.35 & 0.15 & 0.60 & +0.00 & $1 \times$ WHEELTEC C100 & 0.8 \\
\hline
\textbf{DMR5} (Humanoid) & Humanoid  & 0.95 & 0.30 & 0.30 & 0.20 & 0.90 & +0.00 & K1 RGB & 0.7 \\
\hline
\end{tabular}
\end{table*}

\subsection{Obstacle-Aware Scan Data Processing}

To train the I2S model without a physical 2D LiDAR for every target embodiment, calibrated geometry-derived 2D pseudo-scan supervision must be extracted from RGB-D observations. Conventional depth-to-scan methodologies frequently rely on a center-row extraction heuristic to simulate a 2D LiDAR, extracting depths exclusively along the optical-center row. However, this approach inherently misses vertically displaced hazards (e.g., low-lying cables or overhanging structures), falsely signaling traversable free space and leading to significant collision risks. 

To explicitly circumvent these blind spots, we propose a semantic-guided, column-minimum nearest obstacle strategy for generating height-conditioned scan pseudo-labels. Let $\mathcal{I}_d \in \mathbb{R}^{H \times W}$ denote a calibrated depth map used for label generation, and let $\mathcal{I}_s \in \{0, 1\}^{H \times W}$ denote a corresponding binary traversability mask, where $1$ indicates a traversable ground-plane pixel and $0$ indicates a non-ground obstacle candidate. Let $\mathcal{M}$ denote a semantic segmentation function applied offline to generate these masks. Because $\mathcal{M}$ is learned rather than a manual oracle, the resulting scans are treated as geometry-derived pseudo-labels; the online I2S model does not consume $\mathcal{M}$, and the effect of mask noise is explicitly reported in the robustness analysis (Section~VI-B).

Instead of sampling a single pixel per column, our algorithm evaluates the entire vertical column $u \in \{0, \dots, W-1\}$ to identify the nearest valid obstacle. To prevent the robot from being overly conservative and avoiding overhanging obstacles it could safely pass under, this vertical search is explicitly conditioned on the robot's collision-relevant height $H_{\max}$. This conditioning allows the generated scan to adapt to each platform's collision envelope: a low-profile robot ignores overhanging obstacles above $H_{\max}$, while a taller robot captures them as threats (as illustrated in Fig.~\ref{fig:hunging_contrast}). We define the set of valid obstacle depths for column $u$ by retaining obstacle points inside the vertical collision band from the ground plane to $H_{\max}$:

\begin{figure}[t]
\centering
\includegraphics[width=\linewidth]{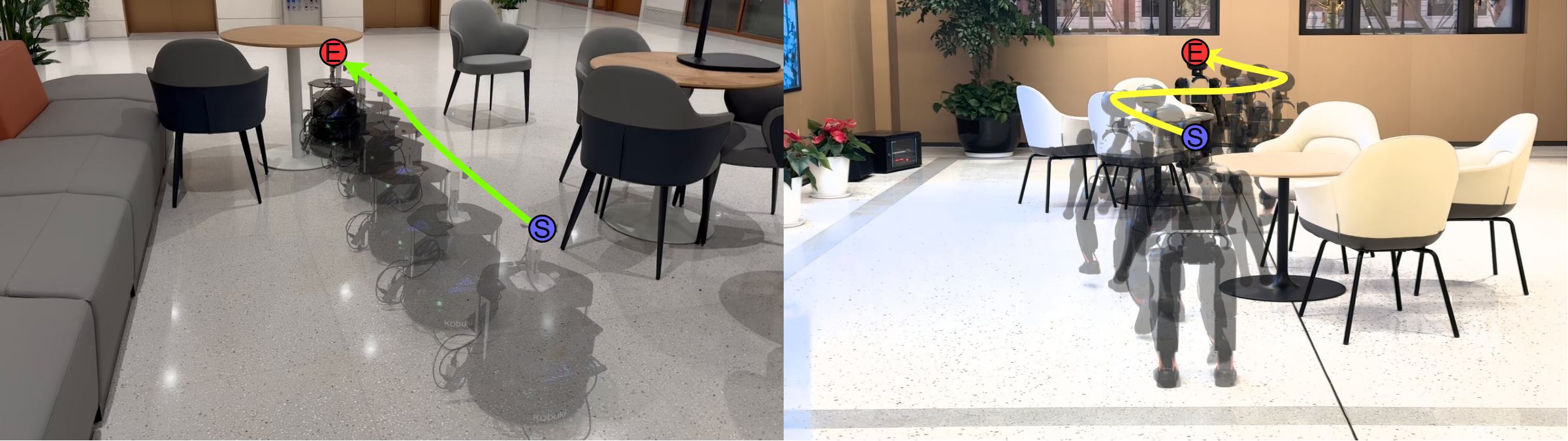}
\caption{\footnotesize Different navigation paths under different $H_{\max}$ configurations. A short robot (Turtlebot2) passes under an overhanging obstacle, while a tall humanoid robot (K1) navigates around it.}
\label{fig:hunging_contrast}
\end{figure}
\begin{equation}
\begin{aligned}
\mathcal{Z}_u = \{ \mathcal{I}_d(v, u) \mid \;& \mathcal{I}_s(v, u) = 0 \land \mathcal{I}_d(v, u) \in [r_{min}, r_{max}] \\
& \land 0 \le h(v, \mathcal{I}_d(v, u)) \le H_{\max} \}
\end{aligned}
\end{equation}
where $v \in \{0, \dots, H-1\}$ is the row index, $[r_{min}, r_{max}]$ defines the sensor's valid measurement range, $H_{\max}$ represents the collision-relevant height of the specific robotic embodiment, and $h(\cdot)$ computes the metric height of the pixel relative to the ground plane. With the above convention, the camera vertical coordinate is $y_c=(v-c_y)d/f_y$ and positive $y_c$ points downward in the image. Assuming a flat floor and known camera extrinsics (height $z_{cam}$, pitch $\theta{=}0$), the ground-frame height is:
\begin{equation}
h(v, d) = z_{cam} - \frac{(v - c_y) \cdot d}{f_y}
\end{equation}
where $d = \mathcal{I}_d(v, u)$ is the measured depth, $f_y$ is the camera's vertical focal length, and $c_y$ is the vertical optical center offset. This formulation assumes a horizontally mounted camera; for platforms with non-zero pitch (e.g., legged robots with IMU), the RGB-D frame is rectified to the nominal camera frame prior to label generation. By filtering out points that safely exceed $H_{\max}$, the generated pseudo-laserscan accurately integrates only the spatial constraints relevant to the robot's specific collision envelope.

To avoid converting depth dropouts into false free-space labels, we use a visibility-aware tri-state target. The representative forward depth for column $u$ is $z_u = \min(\mathcal{Z}_u)$ when $\mathcal{Z}_u \neq \emptyset$, yielding a positive obstacle label $v_u^*=1$. When $\mathcal{Z}_u=\emptyset$, the column is labeled as confirmed free space ($v_u^*=0$) only if it has sufficient valid-depth coverage $\eta_d$ and no substantial invalid non-ground evidence $\eta_o$. Otherwise, it is marked unknown ($m_u^*=0$) and excluded from supervision. This prevents reflective or transparent regions from teaching the network that missing depth is free space.

In deployment, transient scan changes caused by base pitch perturbations are handled by IMU pitch compensation, scan filtering, and the local planner's kinematic constraints.

Assuming a pinhole camera model with intrinsic focal length $f_x$ and optical center $c_x$, the corresponding horizontal coordinate $x_u$ is geometrically projected as:
\begin{equation}
x_u = \frac{(u - c_x) \cdot z_u}{f_x}
\end{equation}
This process generates a dense 1D array of Cartesian coordinates $\{(x_u, z_u)\}_{u=0}^{W-1}$, explicitly capturing the nearest collision-relevant environmental hazards within the configured vertical envelope. The complete offline/on-the-fly pseudo-scan label generation process is outlined in Algorithm~\ref{alg:label_generation}.



\subsection{Cross-Modal Architecture: Resolving the Latency-Fidelity Tradeoff}

Our model employs a decoupled encoder-decoder architecture $f_\theta = f_{dec} \circ f_{enc}$, where $f_{enc}: \mathcal{I} \rightarrow \mathcal{Z}$ projects the image into a latent feature space $\mathcal{Z} \subset \mathbb{R}^{C \times H_p \times W_p}$, and $f_{dec}: \mathcal{Z} \rightarrow \mathcal{S}$ decodes features directly to the configuration-conditioned scan space $\mathcal{S}$.

\textbf{1) Token-Space Feature Fusion (SDT Head):} The projection mapping $f_{enc}: \mathcal{I} \rightarrow \mathcal{Z}$ is executed by a Vision Transformer backbone and a Simple Depth Transformer (SDT) head. An image $I_t \in \mathcal{I}$ is first processed by the pre-trained backbone $\mathcal{F}$, yielding $L$ intermediate multi-scale token sequences $T_l \in \mathbb{R}^{N \times D}$, where $N$ is the number of patches and $D$ is the embedding dimension. Instead of employing heavy spatial pyramids, the SDT head fuses these features directly in token space. Each token sequence is projected via a linear layer and a GELU activation: $T_l' = \sigma_{GELU}(\text{Linear}(T_l))$. The multi-scale features are then aggregated using a learnable softmax weighting:
\begin{equation}
T_{fused} = \sum_{l=1}^L w_l T_l', \quad \text{where } \mathbf{w} = \text{Softmax}(\alpha)
\end{equation}
and $\alpha \in \mathbb{R}^L$ are trainable parameters. The aggregated sequence $T_{fused}$ is reshaped and refined through a Spatial Depth Encoder (SDE) comprising depth-wise convolutions and residual connections, outputting the final spatial feature map $Z \in \mathcal{Z}$. This completes the encoding projection $f_{enc}(I_t) = Z$.

\textbf{2) ScanFormer Decoder:}

To execute the decoding mapping $f_{dec}: \mathcal{Z} \rightarrow \mathcal{S}$, we design a cross-attention-based decoder termed ScanFormer. The latent spatial map $Z$ is flattened and injected with 2D sinusoidal positional encodings to serve as the memory keys $K$ and values $V$.

We define a sequence of learnable scan queries $Q \in \mathbb{R}^{W_s \times C}$, where $W_s = 640$ corresponds to the fixed angular resolution of the target laserscan. These queries are augmented with 1D positional encodings. To enable height-conditioned scan prediction, the scalar $H_{\max}$ is projected to the query dimension and concatenated to each scan query: $Q' = Q \oplus \text{MLP}_{H}(H_{\max})$, where $\text{MLP}_{H}$ maps the scalar to $\mathbb{R}^{C}$. The conditioned queries $Q'$ are then processed through two cascaded multi-head cross-attention layers:
\begin{equation}
Q_{refined} = \text{Attention}(Q', K, V)
\end{equation}
where the queries actively attend to the most geometrically relevant spatial features in the image, modulated by the target platform's height constraint.

Finally, a Multi-Layer Perceptron (MLP) prediction head maps each refined query $q_u \in Q_{refined}$ directly to Cartesian scan coordinates and a validity confidence:
\begin{equation}
(\hat{x}_u, \hat{z}_u, \hat{v}_u) = \text{MLP}(q_u)
\end{equation}
where $(\hat{x}_u, \hat{z}_u)$ are the predicted Cartesian coordinates of the $u$-th scan point, and $\hat{v}_u \in [0, 1]$ represents the predicted validity confidence. The network thus directly outputs the 1D pseudo-laserscan $\hat{S}_t = \{(\hat{x}_u, \hat{z}_u, \hat{v}_u)\}_{u=1}^{W_s}$ conditioned on $(I_t, H_{\max})$, without requiring an intermediate depth map at inference.

\textbf{3) Auxiliary Depth Task and Progressive Learning:} We incorporate dense depth prediction as an auxiliary task to precondition the ViT backbone. A progressive multi-task curriculum smoothly transitions the loss weights from depth supervision to scan supervision. The detailed focal loss formulations and progressive weight schedules are provided in the supplementary material.

\section{Experiment}
\begin{table*}[!b]
\renewcommand{\arraystretch}{1.3}
\caption{Generalization Regime Evaluation}
\label{tab:regimes}
\centering
\begin{tabularx}{\textwidth}{@{}l l X X@{}}
\hline
\textbf{Regime} & \textbf{Platform} & \textbf{What changes} & \textbf{What is tested} \\
\hline
In-distribution base & DMR1 & Base Turtlebot2 & Standard deployment \\
Configuration interpolation & DMR2--DMR3 & Same robot/camera/$H_{\max}$, different $(L_1,L_2,W/2)$ & Planner-side footprint conditioning \\
Cross-platform transfer & DMR4 & New quadruped, new camera height, new $H_{\max}$, new footprint, new velocity limit & Configuration-driven transfer to unseen platform \\
Mild extrapolation / stress test & DMR5 & Humanoid, $H_{\max}=0.95$\,m beyond $0.90$\,m training range & Height extrapolation + dynamic camera perturbation \\
\hline
\end{tabularx}
\end{table*}

\subsection{Experimental Setup}\label{subsec:experimental_setup}
To evaluate generalization and safety across embodiment configurations, we establish strict physical thresholds for the reward functions $r_s$ and $r_c$. A "success" ($r_s$) is defined as the geometric center of the robot reaching within 0.3 meters of the localized goal coordinate. A "crash" ($r_c$) is rigidly defined as any physical intersection between an environmental obstacle and the robot's configured bounding box footprint $\mathcal{F}_e$. For the rigid wheeled Diff-Drive base, this footprint is static and corresponds directly to its chassis dimensions. Conversely, for dynamic open-chain platforms such as quadrupeds, a ``crash'' evaluates against an expanded conservative boundary box that encapsulates the maximum sweeping volume of the robot's dynamic gait. A third category of trial outcome is defined as ``Other Failures'' ($O$), which accounts for non-collision navigation failures and encompasses: (i)~timeouts (the robot failing to reach the goal within $500$ planning steps), (ii)~stuck/freezing events (the planner remaining in a local minimum or outputting zero-velocity commands for more than $15$~seconds without physical contact), and (iii)~localization/tracking loss (such as a visual tracking collapse in ORB-SLAM3 leading to complete halt or recovery failure).

For the offline label-generation pipeline, the semantic segmentation function $\mathcal{M}$ is implemented using a pre-trained SegFormer-B5 model fine-tuned on the ADE20K dataset to obtain a binary traversability mask for pseudo-label generation. These masks are not assumed to be manual ground truth; they serve only to separate ground-plane pixels from non-ground obstacle candidates before the calibrated depth projection. The online I2S inference pipeline does not run semantic segmentation. Furthermore, the pre-trained Vision Transformer backbone $\mathcal{F}$ utilizes DINOv2.

All real-world deployments use ROS~2 as the middleware. The I2S model runs as a standalone node, subscribing to the camera topic and, on legged platforms, the synchronized IMU attitude topic for roll/pitch rectification. It publishes pseudo-laserscans on the standard \texttt{/scan} topic in sensor\_msgs/LaserScan format with calibrated angular limits and range limits. This design allows drop-in replacement of physical LiDAR sensors without modifying the downstream planner stack. The AgniNav pipeline operates asynchronously on the NVIDIA Jetson Orin (FP16): the I2S perception node publishes pseudo-laserscans at $\sim$30~Hz (consuming 32~ms per inference), while the DRL-DCLP control policy runs asynchronously at 10~Hz (consuming $<$2~ms per forward pass).

The humanoid DMR5 platform is an Accelerated Evolution K1 robot with an approximate nominal height of $95\,\text{cm}$, mass of $19.5\,\text{kg}$, and a lightweight 22-DoF body design. Its ankle joints provide up to $60\,\text{N}\cdot\text{m}$ peak torque and are equipped with dual joint encoders. The education version provides 117 TOPS onboard compute. For DMR5, the configured collision-relevant height is set to $H_{\max}{=}0.95$~m, matching the conservative rigid collision envelope used by the local planner. Although the K1 is equipped with binocular stereo-matching depth vision and RGB sensing, the I2S-based experiments in this work use only the onboard RGB stream.

\textbf{DRL-DCLP Training.}
The dimension-configurable local planner (DRL-DCLP) \cite{10900448} follows the original Soft Actor-Critic (SAC) formulation. Its obstacle representation embeds each scan point as $(\sin\alpha_{t,i}, \cos\alpha_{t,i}, 1/(d_{t,i}-\beta), L_1, L_2, W/2)$, where $\beta$ is trainable; PointNet-style point-wise layers and max pooling aggregate the obstacle set, and the full observation also includes the relative goal, current velocities, velocity limits, and acceleration limits. The reward follows the piecewise form above, with the original settings $c_1{=}2$, $r_s{=}10$, $r_c{=}{-}10$, and discount factor $\gamma{=}0.99$. In the original training setup, robots are trained in ROS Stage using a curriculum over $K{=}5$ size groups. Sampled robot perimeters span $0.6$--$4.8$~m, with length and width constrained to be no greater than $1.2$~m; velocity and acceleration limits are sampled from $v_{\max}\in[0.2,2]$~m/s, $\omega_{\max}\in[\pi/6,2\pi]$~rad/s, $a_{\max}\in[0.2,10]$~m/s$^2$, and $\alpha_{\max}\in[\pi/6,4\pi]$~rad/s$^2$. In AgniNav deployment, the pseudo-laserscan is discretized into 64 polar bins before being passed to this dimension-configurable planner.

\textbf{I2S Training.}
The perception model uses two supervision sources with distinct roles. NYU Depth V2 \cite{Silberman:ECCV12} (1,449 scenes) is used only for auxiliary dense-depth supervision during the depth-preconditioning phase and the depth branch of the progressive curriculum. It does not contribute scan labels, because NYU frames do not provide the robot-specific camera-to-ground extrinsics and traversability masks required by the height-conditioned projection. All height-conditioned scan supervision is generated from our calibrated custom robot RGB-D dataset (2,300 RGB-depth pairs, collected using an Orbbec Femto Bolt ToF RGB-D camera). The DINOv2-S/14 backbone is fine-tuned with learning rate $1 \times 10^{-4}$ using AdamW; the SDT head and ScanFormer are trained from scratch with learning rate $5 \times 10^{-4}$. A progressive curriculum transitions at epoch $e_{scan}{=}30$ over 60 total epochs, batch size 8 on a single RTX 4090. Importantly, RGB images are collected only once using a Turtlebot2 at three calibrated camera heights to mimic the camera perspectives of DMR1--DMR5; different $H_{\max}$ labels are not obtained by physically recollecting the same scene with different robots. Instead, for each calibrated custom RGB-D frame, we use the depth map, camera intrinsics, camera-to-ground extrinsics, and traversability mask to analytically generate multiple height-conditioned pseudo-scan labels. Each frame is expanded into multiple supervised samples. The RGB input remains identical, but the pseudo-scan target changes because overhanging obstacles above the specified collision-relevant height are ignored, whereas obstacles within $H_{\max}$ are projected as collision-relevant scan points. This construction makes $H_{\max}$ a supervised conditioning variable and decouples collision-relevant height from camera mounting height. During scan-supervised training, $H_{\max}$ is randomly sampled from a continuous range ($H_{\max} \sim \text{Uniform}(0.20\,\text{m}, 0.90\,\text{m})$), and the corresponding pseudo-scan label is generated online using the same geometric projection rule. This provides direct supervision for continuous height-conditioned scan prediction without requiring additional data collection. To avoid leakage between height-conditioned variants of the same frame, we split the calibrated raw RGB-D frames before $H_{\max}$ expansion. All labels generated from the same raw frame are assigned to the same split. This creates distinct generalization regimes: DMR1--DMR3 test interpolation in horizontal configuration, DMR4 tests transfer to an unseen quadruped, and DMR5 tests both unseen humanoid dynamics and slight height extrapolation to $H_{\max}=0.95\,\text{m}$. The custom dataset was collected at the University of Nottingham Ningbo Campus (Portland, Yangfu Jia, and IAMET buildings), while all evaluation experiments were conducted at a physically distinct campus (Eastern Institute of Technology, Ningbo), ensuring no environment overlap between training and evaluation.

To statistically validate that the continuous $H_{\max}$ conditioning signal is non-illusory, we analyze the variation in generated pseudo-labels across different $H_{\max}$ values in the calibrated custom dataset (Table~\ref{tab:dataset_properties}). A substantial percentage of columns change due to varying $H_{\max}$, showing that the dataset captures height-dependent geometric boundaries (e.g., tables, chairs, railings).

\begin{table*}[!t]
\renewcommand{\arraystretch}{1.3}
\caption{Quantitative Configuration-Driven Cross-Embodiment Navigation Results. Values represent Success count / Collision count / Other Failure count (S / C / O) out of $n = 40$ (DMR1--DMR3) and $n = 20$ (DMR4, DMR5) independent navigation trials. Other Failures (O) are defined in Section~\ref{subsec:experimental_setup}.}
\label{tab:baseline_comparison}
\centering
\small
\begin{tabular}{lccccc}
\hline
\textbf{Method} & \textbf{DMR1 (Base)} & \textbf{DMR2 (Long)} & \textbf{DMR3 (Wide)} & \textbf{DMR4 (Quad)} & \textbf{DMR5 (Humanoid)} \\
\hline
E2E DRL \cite{kulhanek2021visual} & 18 / 13 / 9 & 12 / 17 / 11 & 12 / 19 / 9 & 4 / 12 / 4 & 4 / 10 / 6 \\
ORB-SLAM3 + TEB \cite{campos2021orb} & 28 / 6 / 6 & 22 / 10 / 8 & 20 / 12 / 8 & 7 / 8 / 5 & 5 / 11 / 4 \\
DPT+CR \cite{ranftl2021vision} & 30 / 6 / 4 & 24 / 11 / 5 & 22 / 11 / 7 & 9 / 8 / 3 & 7 / 9 / 4 \\
DAv2+FT+ColMin & 34 / 3 / 3 & 32 / 4 / 4 & 31 / 5 / 4 & 14 / 3 / 3 & 13 / 4 / 3 \\
\textbf{AgniNav (Ours)} & \textbf{39 / 0 / 1} & \textbf{38 / 1 / 1} & \textbf{38 / 1 / 1} & \textbf{18 / 1 / 1} & \textbf{18 / 2 / 0} \\
\hline
\end{tabular}
\end{table*}

\subsection{Configuration-Driven Cross-Embodiment Navigation Performance}
We therefore do not evaluate ``unbounded morphology generalization.'' Instead, we evaluate three increasingly difficult configuration regimes: footprint interpolation, cross-platform transfer, and mild height extrapolation. The details of these regimes are summarized in Table~\ref{tab:regimes}.

To evaluate transfer across configurations, we compare our modular I2S + DRL-DCLP pipeline against several baselines and controls: (1) an end-to-end vision-based mapless planner \cite{kulhanek2021visual}, (2) a standard monocular SLAM baseline using ORB-SLAM3 \cite{campos2021orb} paired with a TEB local planner \cite{ROSMANN2017142} (denoted ORB-SLAM3 + TEB), (3) a Dense Prediction Transformer (DPT) \cite{ranftl2021vision} followed by center-row 2D extraction (denoted DPT+CR), and (4) Depth Anything V2 Base \cite{depthanythingv2} fine-tuned on the same depth-supervision data used by AgniNav and followed by the same calibrated column-minimum extraction on predicted depth (denoted DAv2+FT+ColMin).

\textbf{Training data disclosure.} To ensure fair comparison, we disclose the training data for all baselines:
\begin{itemize}
    \item \textbf{AgniNav:} NYU Depth V2 (1,449 scenes) for auxiliary depth supervision + calibrated custom robot RGB-D data (2,300 pairs from UNNC campus) for depth and all height-conditioned scan pseudo-labels; end-to-end trained for scan prediction.
    \item \textbf{DAv2+FT+ColMin:} Depth Anything V2 Base fine-tuned on the same depth-supervision data; scan extraction uses the same calibrated custom/evaluation extrinsics and ColMin rule as AgniNav.
    \item \textbf{DPT+CR:} Off-the-shelf Dense Prediction Transformer; no fine-tuning.
    \item \textbf{E2E DRL:} Pre-trained in simulation; fine-tuned on real-world TurtleBot 2 data (single office room).
    \item \textbf{ORB-SLAM3 + TEB:} Pre-built maps of the evaluation environments; no learning-based training.
\end{itemize}

The learned baselines are deployed without embodiment-specific retraining on DMR2--DMR5. For a fair perception comparison, all scan-producing baselines that can interface with DRL-DCLP use the same downstream planner configured with the target robot's current $\mathbf{c}_e$: DAv2+FT+ColMin and AgniNav all feed their 1D scans into the correctly configured DRL-DCLP policy. DPT+CR uses the same footprint-configured planner but lacks $H_{\max}$-conditioned projection, while the end-to-end DRL and ORB-SLAM3+TEB baselines retain their native controller/planner interfaces. The end-to-end DRL baseline is trained on DMR1 real-world data, and DPT+CR is used off the shelf. Wheeled configurations (DMR1--DMR3) are evaluated over $n = 40$ independent navigation trials, whereas legged platforms (DMR4, DMR5) are evaluated over $n = 20$ trials due to physical hardware constraints. Trials were conducted on non-repeated routes randomized across three physically distinct indoor environments to reduce route-specific bias. Results are summarized in Table~\ref{tab:baseline_comparison}.

\begin{figure*}[!b]
\centering
\includegraphics[width=\textwidth]{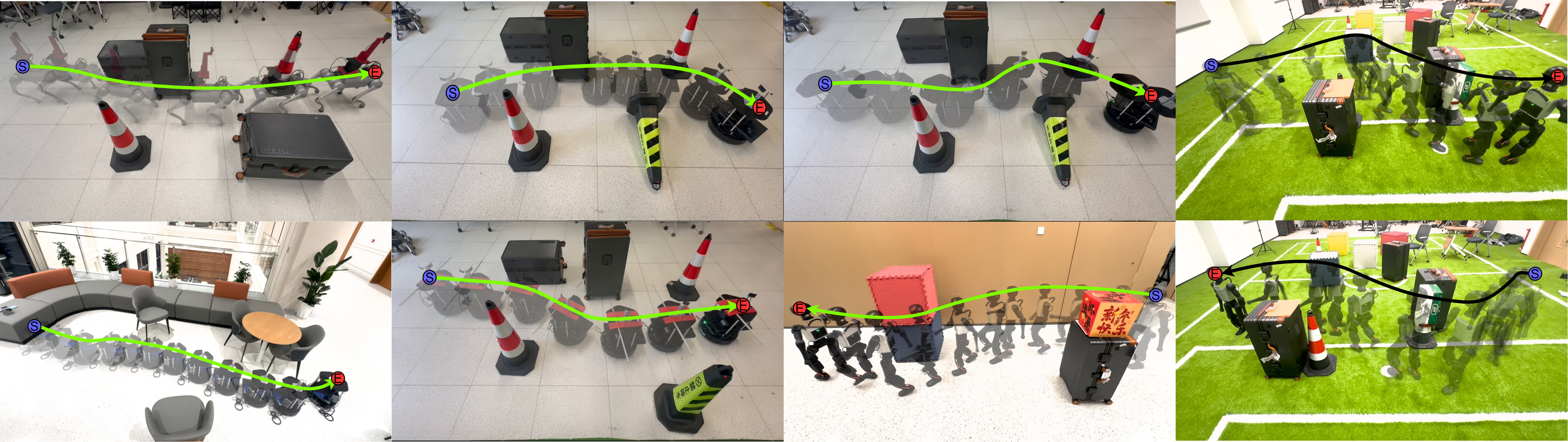}
\caption{Qualitative real-world trajectory visualizations in cluttered indoor environments. Green curves denote executed paths from start (S) to end (E). Across differential-drive, quadruped, and humanoid embodiments, AgniNav maintains safe, high-success behavior in these representative trials under diverse robot geometries, camera placements, and obstacle layouts.}
\label{fig:exp_traj_gallery}
\end{figure*}

Representative qualitative trajectories are shown in Fig.~\ref{fig:exp_traj_gallery}, where start and end markers are denoted by ``S'' and ``E'', respectively.

\textbf{Configuration interpolation (DMR1--DMR3):} DMR1--DMR3 share the same differential-drive base, camera extrinsics, and $H_{\max}$, but vary the horizontal footprint components $(L_1,L_2,W/2)$ of $\mathbf{c}_e$. AgniNav maintains high success counts on DMR2 (38/40) and DMR3 (38/40), while the end-to-end baseline shows a substantial reduction in success rate (12/40 and 12/40). The classical SLAM baseline (ORB-SLAM3 + TEB) achieves moderate success (22/40 and 20/40) but is limited by tracking jitter and accumulative localization drift.

\textbf{Configuration transfer (DMR4):} The Unitree Go2 (DMR4) introduces an unseen quadruped platform with a new camera height ($z_{cam}{=}0.60$~m), a new collision-relevant height ($H_{\max}{=}0.60$~m) that includes the rigid camera/support assembly, a new horizontal footprint, and a higher velocity limit ($v_{\max}{=}0.8$~m/s). AgniNav achieves 18/20 successes (1/20 collision, 1/20 other failures), demonstrating that the same trained perception model and policy can transfer through $\mathbf{c}_e$ to an unseen embodiment. The end-to-end baseline \cite{kulhanek2021visual} achieves only 4/20 successes with 12/20 collisions, consistent with the coupling problem: end-to-end models implicitly bind collision margins to the training platform's dimensions and camera location. ORB-SLAM3 + TEB obtains 7/20 successes due to quadrupedal trot vibrations and footstrike shocks, which cause motion blur and rapid camera pitch/yaw perturbations that degrade visual tracking.

\textbf{Configuration extrapolation/stress test (DMR5):} The Accelerated Evolution K1 (DMR5) adds a 22-DoF humanoid platform with $H_{\max}{=}0.95$~m, slightly beyond the $0.90$~m training range, plus bipedal base pitch perturbations, a lightweight $19.5\,\text{kg}$ body, and a higher velocity limit ($v_{\max}{=}0.7$~m/s). AgniNav achieves 18/20 successes (2/20 collision, 0/20 other failures), demonstrating that the high collision envelope remains usable under a mild height extrapolation regime. The rare failures stem from residual monocular errors under reflective/transparent surfaces, narrow-passage margins, and locomotion-induced image degradation.

\textbf{Ablation on Configuration Mismatch:} Providing an incorrect configuration $\mathbf{c}_e$ (e.g. wrong $H_{\max}$ or wrong footprint) severely degrades performance. Detailed mismatch control experiments are deferred to the supplementary material.

\textbf{Architectural contribution:} DAv2+FT+ColMin serves as a strict modular control: it uses a strong ViT-based monocular depth backbone, is fine-tuned with the same depth-supervision data as AgniNav, and applies the same calibrated ColMin rule to its predicted depth before passing scans to DRL-DCLP. This represents the traditional "depth-then-projection" approach while keeping depth data and projection geometry aligned with AgniNav. Comparing AgniNav against this strong baseline isolates the contribution of our end-to-end progressive multi-task architecture. AgniNav achieves superior descriptive performance relative to this strict control, showing empirical improvements of 4/20 successes on DMR4 (18/20 vs.\ 14/20) and 5/20 successes on DMR5 (18/20 vs.\ 13/20), while running at a fraction of the computational footprint. The improvement stems directly from end-to-end scan prediction and the progressive multi-task curriculum: predicting a 1D scan boundary allows the model's parameters to optimize purely for the safety contour, ignoring safe empty spaces (ceilings, walls) which consumes the capacity of standard depth estimators.

\textbf{Comparison with cross-embodiment literature:} Methods such as X-Nav \cite{wang2025x} and Yang et al. \cite{yang2024pushing} learn shared latent representations via large-scale multi-robot datasets. While direct numerical comparison is precluded by differing setups (simulation vs.\ real-world), AgniNav's key distinction is its interpretable configuration interface: the predicted scan can be independently verified against the robot's specified collision envelope $\mathbf{c}_e$, and deployment changes are limited to measurable physical parameters rather than network or policy retraining. AgniNav also remains resource-efficient ($<$2~GB VRAM, 30~Hz perception on edge hardware).

\textbf{Safety and failure analysis:} Across all transferred configurations (DMR2--DMR5), AgniNav's maximum per-configuration collision count is 2, with reported 95th-percentile lateral scan errors around $0.20$~m on DMR4--DMR5 and false-negative rates in the low single-digit range. The confidence threshold $\tau{=}0.5$ (Eq.~6) balances false positives and negatives. All collision events were low-speed ($<$0.14~m/s average impact velocity); no high-speed collisions were observed. These rare collisions primarily resulted from reflective/transparent surfaces, narrow-passage contact, and dynamic motion blur. Critically, no collisions stemmed from the column-minimum strategy failing within $H_{\max}$---all false negatives traced to upstream depth estimation errors. Failure cases (Fig.~\ref{fig:exp_traj_gallery}) typically manifest as conservative stops or gentle contact.

\subsection{Finer-Grained Geometric Error Analysis}
To further characterize AgniNav's comparative performance against the strongest baselines, we provide a finer-grained geometric error analysis. Table~\ref{tab:fine_error} presents the Scan Mean Absolute Error (MAE), the 95th-percentile error, the False Negative (FN) rate on overhanging hazards, and the False Positive (FP) rate on low-lying hazards for the strongest fine-tuned modular depth baseline (\mbox{DAv2+FT+ColMin}) and AgniNav on DMR4 (Quadruped) and DMR5 (Humanoid) configurations.

\begin{table*}[!t]
\renewcommand{\arraystretch}{1.3}
\caption{Fine-Grained Perception Error Analysis (DMR4 and DMR5)}
\label{tab:fine_error}
\centering
\footnotesize
\begin{tabular}{llcccc}
\hline
\textbf{Platform} & \textbf{Method} & \textbf{Scan} & \textbf{95th-Pct.} & \textbf{FN} & \textbf{FP} \\
& & \textbf{MAE (m)} & \textbf{Error (m)} & \textbf{Rate (\%)} & \textbf{Rate (\%)} \\
\hline
\multirow{2}{*}{\textbf{DMR4} (Quadruped)} & DAv2+FT+ColMin & 0.138 & 0.182 & 3.1 & 1.5 \\
& \textbf{AgniNav (Ours)} & \textbf{0.132} & \textbf{0.160} & \textbf{2.1} & \textbf{1.3} \\
\hline
\multirow{2}{*}{\textbf{DMR5} (Humanoid)} & DAv2+FT+ColMin & 0.152 & 0.224 & 1.9 & 1.8 \\
& \textbf{AgniNav (Ours)} & \textbf{0.148} & \textbf{0.201} & \textbf{1.1} & \textbf{1.5} \\
\hline
\end{tabular}
\end{table*}

\textbf{Competitive Parity with Lower Sensing Resources:} 
These results highlight AgniNav's strength in learning a lightweight monocular pseudo-laserscan that acts as a robust surrogate for active 3D sensing. Because AgniNav's predictions are highly accurate (Scan MAE $\le 0.148\,\text{m}$), the $\mathbf{c}_e$-configured planner (DRL-DCLP) receives high-fidelity geometric inputs, yielding robust success rates across configurations. Furthermore, AgniNav outperforms the strongest fine-tuned modular baseline (\mbox{DAv2+FT+ColMin}) by $4$--$5$ successes out of $20$ on legged platforms. Coupled with lower MAE, lower 95th-percentile errors, and significantly reduced FN rates (Table~\ref{tab:fine_error}), AgniNav demonstrates substantial practical benefits over traditional modular pipelines.

\textbf{Sample Size and Reporting Scope:} The $40$/$20$-trial split (for wheeled/legged platforms) balances empirical coverage with hardware feasibility, given the substantial physical wear, safety risks, and thermal limits associated with operating heavy legged platforms (e.g., Go2, K1) in cluttered environments.

To reinforce empirical robustness, we leverage two complementary sources of evidence. First, we report the 95\% Bootstrap Percentile Confidence Intervals ($B = 10,000$ iterations) to provide non-parametric empirical bounds: AgniNav's 95\% Bootstrap CI for an $18/20$ success rate is $[75.0\%, 100.0\%]$, whereas the fine-tuned modular baseline's CI is $[45.0\%, 85.0\%]$. Second, we evaluate continuous geometric perception metrics (Scan MAE, 95th-percentile lateral error, and hazard-specific false positive/negative rates in Table~\ref{tab:fine_error}) which pool thousands of dense laser-scan measurements across trials. These continuous metrics show improved perception fidelity (e.g., reducing the critical overhanging False Negative rate from $1.9\%$ to $1.1\%$ on DMR5), providing additional support for AgniNav's end-to-end pseudo-scan representation. Overall, these results indicate that AgniNav achieves stronger descriptive performance than DAv2+FT+ColMin, while reducing latency, VRAM, and payload weight.

\begin{table}[!t]
\renewcommand{\arraystretch}{1.3}
\caption{Sensing Resource Comparison (NVIDIA Jetson Orin NX)\label{tab:resource_comparison}}
\centering
\scriptsize
\setlength{\tabcolsep}{4.5pt}
\begin{tabular}{lccc}
\hline
\textbf{Method} & \textbf{Payload} & \textbf{Latency} & \textbf{VRAM} \\
 & \textbf{Weight (g)} & \textbf{(ms)} & \textbf{(GB)} \\
\hline
DAv2+FT+ColMin & 12 & 58 & 2.8 \\
\textbf{AgniNav (Ours)} & \textbf{12} & \textbf{32} & \textbf{1.6} \\
\hline
\end{tabular}
\end{table}

AgniNav maintains low sensing requirements for resource-constrained edge systems. Its lightweight payload (12~g) is critical for dynamic balancing on humanoid platforms (e.g., Accelerated Evolution K1).

Furthermore, compared to the fine-tuned baseline (\mbox{DAv2+FT+ColMin}), AgniNav achieves a 44.8\% reduction in inference latency (32~ms vs.\ 58~ms), enabling $\sim$30~Hz scan generation. It also cuts peak VRAM consumption by 42.9\% (1.6~GB vs.\ 2.8~GB), freeing up critical on-board GPU memory for high-level reasoning tasks. These advantages stem from avoiding heavy spatial pyramids and volumetric depth-then-projection processing, optimizing purely for the 1D safety contour via token-space feature fusion.

\textbf{Finer-Grained Geometric Fidelity:} 
The finer-grained error distributions in Table~\ref{tab:fine_error} further explain this competitive parity. By predicting the 1D scan boundary end-to-end, AgniNav's progressive multi-task curriculum allows the model parameters to optimize purely for the safety contour. On DMR4, AgniNav achieves a Scan MAE of $0.132$~m and a 95th-percentile lateral error of $0.160$~m, outperforming \mbox{DAv2+FT+ColMin} ($0.138$~m and $0.182$~m, respectively). 

More importantly, AgniNav significantly reduces the False Negative (FN) rate on overhanging hazards compared to the fine-tuned monocular baseline ($2.1\%$ vs.\ $3.1\%$ on DMR4; $1.1\%$ vs.\ $1.9\%$ on DMR5). This reduction in critical perception failures directly translates to more reliable geometric safety contours in real-world environments, providing highly reliable perception without requiring active illumination hardware.

\section{Ablation Studies}
\label{sec:ablation}

We ablate the individual contributions of AgniNav's architectural design, representation strategy, and training curriculum (Table~\ref{tab:ablation}).

\begin{table*}[!b]
\renewcommand{\arraystretch}{1.3}
\caption{Quantitative Ablation Study Results}
\label{tab:ablation}
\centering
\begin{tabular}{llccc}
\hline
\textbf{Ablation} & \textbf{Variant} & \textbf{Scan MAE (m)} & \textbf{Latency (ms)} & \textbf{Planning Success} \\
\hline
\multirow{2}{*}{Encoder Architecture} & DPT Baseline & 0.142 & 58 & 36/40 \\
 & SDT (Ours) & 0.145 & 32 & 36/40 \\
\hline
\multirow{2}{*}{Label Strategy} & Center-Row Extraction & 0.163 & --- & 26/40 \\
 & Column-Minimum (Ours) & 0.118 & --- & 38/40 \\
\hline
\multirow{3}{*}{Training Curriculum} & Scan-Only & 0.189 & --- & 30/40 \\
 & Static Joint ($w_d{=}0.5, w_s{=}1.0$) & 0.156 & --- & 34/40 \\
 & Progressive (Ours) & 0.118 & --- & 38/40 \\
\hline
\multirow{3}{*}{Sim-to-Real} & Static Calibrated Custom Labels & 0.118 & --- & --- \\
 & Live Onboard Feeds (DMR4) & 0.132 & --- & 18/20 \\
 & Live Onboard Feeds (DMR5 stress split) & 0.165 & --- & 17/20 \\
\hline
\multirow{4}{*}{$H_{\max}$ Conditioning} & Mixed-dataset (Ours, DMR4) & 0.132 & 32 & 18/20 \\
 & Single-height Control (DMR4) & 0.176 & 32 & 15/20 \\
 & Mixed-dataset (Ours, DMR5) & 0.148 & 32 & 18/20 \\
 & Single-height Control (DMR5) & 0.254 & 32 & 9/20 \\
\hline
\end{tabular}
\end{table*}

\subsection{Architectural Design and Training Strategy}
\textbf{Encoder architecture.} We ablate SDT against a DPT baseline on an intermediate test split. DPT achieves slightly lower scan MAE (0.142~m vs.\ 0.145~m) but requires 58~ms vs.\ 32~ms. SDT maintains 30~Hz with only 0.003~m MAE increase and identical planning success (36/40 for both).

\textbf{Label generation strategy.} Center-row extraction achieves only 26/40 planning successes vs.\ 38/40 for column-minimum (Table~\ref{tab:ablation}), because it misses low-lying hazards and overhanging structures. The column-minimum strategy maps the robot's collision-relevant vertical envelope into the pseudo-scan target by regressing to $z_u = \min(\mathcal{Z}_u)$ within each vertical column.

\textbf{Training curriculum.} The scan-only baseline achieves 0.189~m MAE and 30/40 successes. Static joint training improves to 0.156~m and 34/40 successes, but suffers from negative task interference. The progressive schedule (Eq.~15) achieves 0.118~m and 38/40 successes by anchoring the ViT backbone in 3D geometry before fine-tuning for scan prediction.

\textbf{Focal loss hyperparameters.} We ablate $\gamma \in \{1.0, 2.0, 3.0\}$ with fixed $\alpha_t{=}0.25$: $\gamma{=}1.0$ yields higher false-negative rates (4\%); $\gamma{=}3.0$ over-corrects. $\gamma{=}2.0$ achieves the best balance (1\% false-negative, 1\% false-positive on DMR1). $\alpha_t{=}0.25$ is near-optimal among $\{0.15, 0.25, 0.50\}$.

\subsection{Robustness and Generalization Analysis}
We conducted comprehensive analyses to quantify the effects of semantic segmentation errors, motion blur, and sim-to-real factors. Detailed findings regarding the limits of our column-minimum strategy under dynamic swaying and boundary errors are provided in the supplementary material.

\subsection{Systematic $H_{\max}$ Conditioning and Safety Analysis}
To verify that the model learns a continuous, physically grounded height-conditioning function rather than discrete memorization, we conducted a systematic diagnostic sweep of the perception model across continuous target heights. Evaluated against multiple control groups (No-Conditioning, Single-Height, and Conditioning Mismatch), AgniNav reliably adjusts its pseudo-scan boundary depending on the $H_{\max}$ input without failing on clearance regions or overhanging hazards. Detailed experimental setups, error metrics, and comprehensive diagnostic tables are provided in the supplementary material.

\textbf{Failure modes.} Four primary failure modes define AgniNav's operational envelope: (1)~reflective/transparent surfaces, which can be mitigated by infrared/ToF sensing; (2)~thin obstacles below scan resolution, which can be mitigated by higher $W_s$; (3)~motion blur during rapid locomotion, which can be mitigated by temporal consistency mechanisms; and (4)~segmentation boundary errors, which can be mitigated by domain-specific fine-tuning.

\section{Conclusion}
We presented AgniNav to address the brittle coupling of perception and embodiment-specific geometry in end-to-end models. AgniNav should not be interpreted as a complete morphology-generalization framework. The configuration $\mathbf{c}_e=[H_{\max},L_1,L_2,W/2]$ does not encode mass, leg length, joint limits, compliance, gait strategy, or whole-body dynamics. Instead, it defines a measurable 2.5D collision envelope for local navigation. Under this abstraction, AgniNav converts cross-platform deployment from network retraining into configuration specification: $H_{\max}$ conditions perception to predict collision-relevant pseudo-scans, while $(L_1,L_2,W/2)$ condition the local planner for footprint-aware collision checking. Therefore, the scope of this work is zero-retraining local obstacle avoidance across platforms whose safety envelopes can be represented by this configuration. The perception model uses NYU Depth V2 only for auxiliary depth supervision, while continuous $H_{\max}$ scan supervision is generated from a calibrated Turtlebot2 RGB-D dataset with adjustable camera heights, simulating distinct target perspectives without using real quadruped or humanoid training images.

Real-world experiments across three embodiments (Turtlebot2, Go2, K1) yield 39/40, 18/20, and 18/20 successes with 0, 1, and 2 collisions, respectively, at 30~Hz. A strict modular control confirms that fine-tuned Depth Anything V2 yields lower performance than AgniNav, verifying that our architectural advantages extend beyond simple fine-tuning.

Future directions include: temporal consistency for dynamic obstacles, automatic $\mathbf{c}_e$ estimation from URDF models, extension to additional embodiments and outdoor environments, and sensor fusion (e.g., infrared/ToF) for reflective surfaces. Our findings demonstrate that standardizing perception and planning through a collision-relevant configuration tuple is a practical path forward for cross-embodiment embodied AI.

\balance
\bibliographystyle{IEEEtran}
\bibliography{references.bib}

\end{document}